\documentclass{egpubl}
\usepackage{DH2025}%
 
\WsPaper           % uncomment for final version of EG Workshop contribution
\usepackage[T1]{fontenc}
\usepackage{dfadobe}  
\usepackage{subcaption}
\usepackage{float}
\usepackage{multirow}
\usepackage{comment}
\usepackage{siunitx}
\usepackage{tablefootnote}
\usepackage{gensymb}
\usepackage[dvipsnames]{xcolor}
\usepackage{amsmath}
\usepackage{amsfonts} 
\usepackage{tikz}
\newcommand*\circled[1]{\tikz[baseline=(char.base)]{
            \node[shape=circle,draw,inner sep=1pt] (char) {#1};}}
\usepackage[ruled,vlined]{algorithm2e}

\usepackage{cite}  % comment out for biblatex with backend=biber
\usepackage{arydshln} % Pour utiliser \hdashline dans les tableaux
\usepackage{dblfloatfix} % Pour la gestion des flottants en 2 colonnes

\BibtexOrBiblatex
\electronicVersion
\PrintedOrElectronic
\usepackage{egweblnk} 
\title[Identification of Violin Reduction through Contour Lines Classification]%
      {Identification of Violin Reduction via Contour Lines Classification}

\author[Philémon Beghin \& Anne-Emmanuelle Ceulemans \& François Glineur]
{\parbox{\textwidth}{\centering Philémon Beghin$^{1,2}$\orcid{0000-0003-0354-7901}, Anne-Emmanuelle Ceulemans$^{1,3,4}$\orcid{0000-0002-3273-4004} and François Glineur$^{1,2}$\orcid{0000-0002-5890-1093} 
        }
        \\
{\parbox{\textwidth}{\centering $^1$UCLouvain, Louvain-la-Neuve, Belgium\\
         $^2$Institute of Information and Communication Technologies, Electronics and Applied Mathematics (ICTEAM), Louvain-la-Neuve, Belgium\\
         $^3$Institute for the Study of Civilisations, Arts and Letters (INCAL), Louvain-la-Neuve, Belgium\\
         $^4$ Musical Instruments Museum (MIM), Brussels, Belgium
       }
}
}
\begin{document}

% uncomment for using teaser
% \teaser{
%  \includegraphics[width=0.9\linewidth]{eg_new}
%  \centering
%   \caption{New EG Logo}
% \label{fig:teaser}
%}

\maketitle
%-------------------------------------------------------------------------
\begin{abstract}
   The first violins date back to the end of the 16th century in Italy. For around 200 years, these instruments have spread throughout Europe and luthiers of various royal courts, eager to experiment with new techniques, created a highly diverse family of instruments. In an attempt to normalise violins for European orchestras and conservatories, size standards were imposed around 1750. Instruments that fell between two standards were then reduced to a smaller size by luthiers. These reductions have an impact on several characteristics of violins, in particular on the contour lines, i.e. lines of constant altitude as measured from a reference plane between the violin plates, which look more like a ‘U’ for non reduced instruments and a ‘V’ for reduced ones. Those differences between (un)reduced violins have been observed empirically but to our knowledge no quantitative study has been carried out on the subject. \\

In this paper, we aim at developing a tool for classifying violin contour lines in order to distinguish reduced instruments from non reduced instruments. We study a corpus of 25 instruments whose 3D geometric meshes have been acquired via photogrammetry. For each instrument, we sample contour lines at 10-20 levels, regularly spaced every millimetre. Each contour line is fitted with a parabola-like curve (with an equation of the type $y = \alpha |x|^\beta$) depending on two parameters, describing how open ($\beta$) and how vertically stretched ($\alpha$) the curve is. We compute additional features from those parameters, using regressions and counting how many values fall under some threshold. We also deal with outliers and non equal numbers of levels, and eventually obtain a numerical profile for each instrument.\\

We then applied different learning techniques on those profiles to determine whether instruments can be classified solely according to their geometry. We find that distinguishing between reduced and non reduced instruments is feasible to some degree, taking into account that a whole spectrum of more or less transformed violins exists, for which it is more difficult to quantify the reduction. We also find the opening parameter $\beta$ to be the most predictive.
%-------------------------------------------------------------------------
%  ACM CCS 1998
%  (see https://www.acm.org/publications/computing-classification-system/1998)
% \begin{classification} % according to https://www.acm.org/publications/computing-classification-system/1998
% \CCScat{Computer Graphics}{I.3.3}{Picture/Image Generation}{Line and curve generation}
% \end{classification}
%-------------------------------------------------------------------------
%  ACM CCS 2012
   %(see https://www.acm.org/publications/class-2012)
%The tool at \url{http://dl.acm.org/ccs.cfm} can be used to generate
% CCS codes.
%Example:


%\printccsdesc   
\end{abstract}  
%-------------------------------------------------------------------------
\section{Historical context and preliminary observations}

It took around 200 years for the violin family, which dates back to the end of the sixteenth century in Italy, to adopt standard dimensions. Before about 1750, the size of instruments and their names differed from one country to another and from one luthier to another. Once the new standards had been established (namely the violin, viola and cello, which are still the current references), luthiers reduced instruments which stand between two sizes in order to make them fit a smaller size. An important question that arises in musicology nowadays is to determine whether or not an instrument has been reduced, and if so, to quantify how; this issue is also relevant for its cultural and economic aspects. A detailed historical review of violin reduction can be found in \cite{ceulemans2023baroque,zeller2019reconstructing}. It should be noted that the reduction problem only concerns the sound box of the instruments, and not the neck, which has often been changed and replaced over time \cite{sibire1885chelonomie,stowell1990violin,herzog2003quinton}.

\begin{figure}
  \centering
  \begin{subfigure}{0.27\linewidth}
    \centering
    \includegraphics[scale = 0.13]{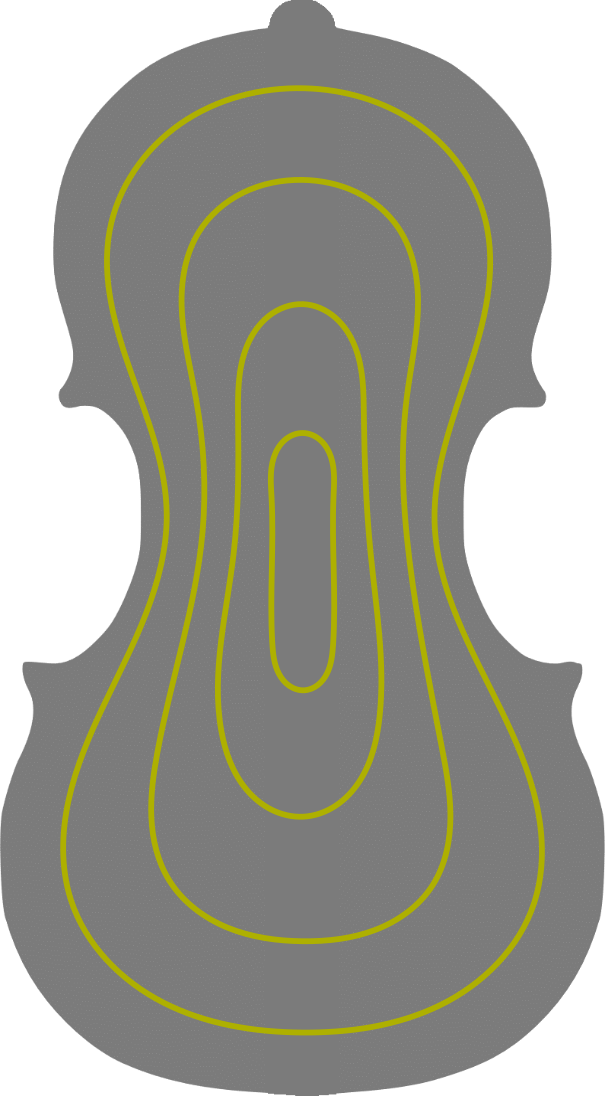}
  \end{subfigure}%
  %\hspace{-0.5cm}% Space between image A and B
  \begin{subfigure}{0.24\linewidth}
    \centering
    \includegraphics[scale = 0.13]{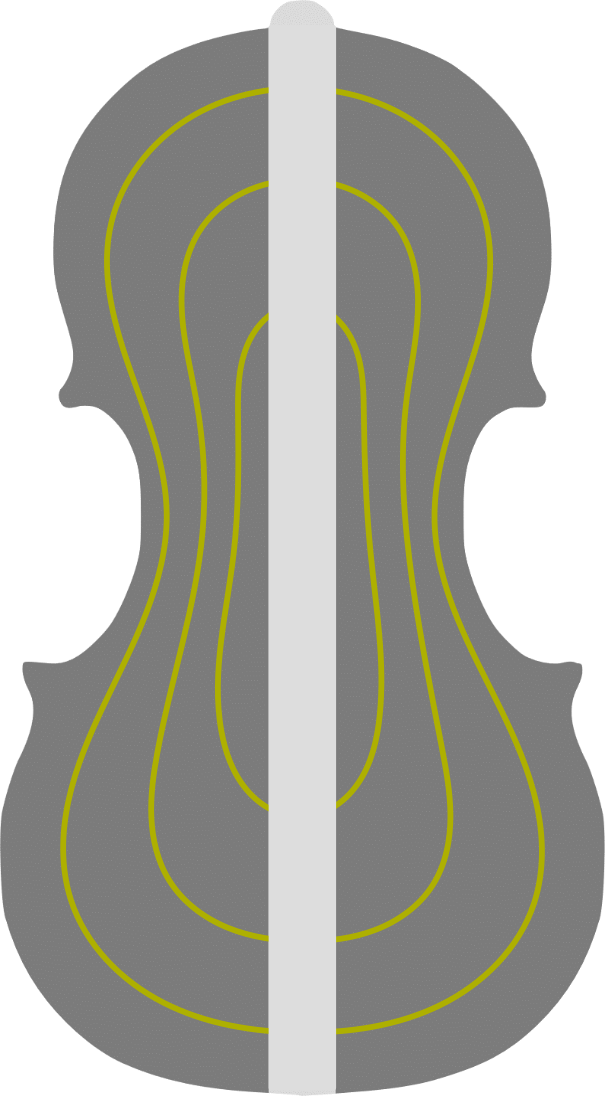}
  \end{subfigure}%
  %\hspace{0.5cm}% Space between image B and C
  \begin{subfigure}{0.27\linewidth}
    \centering
    \includegraphics[scale = 0.2]{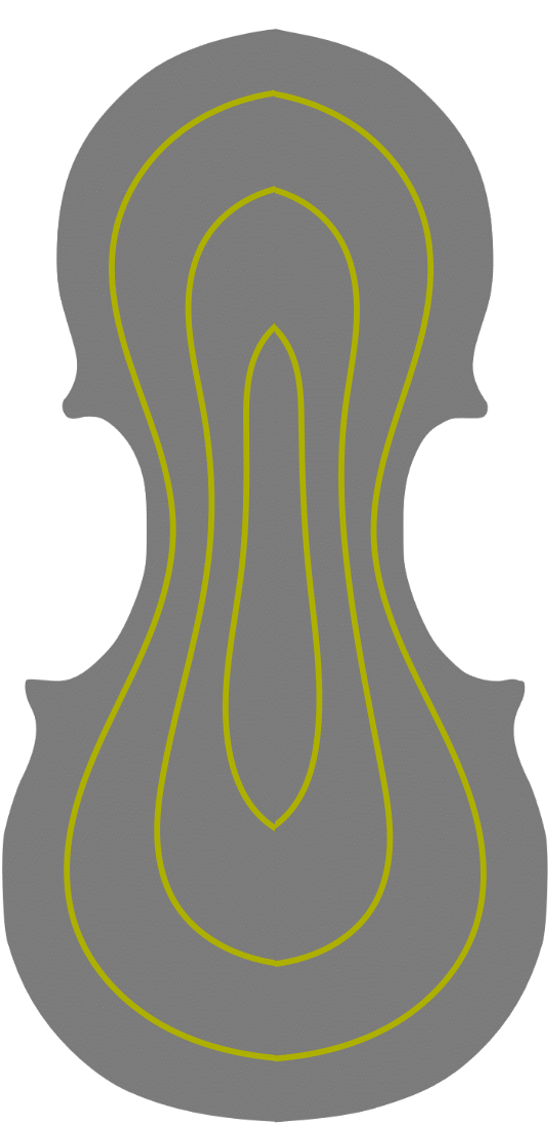}
  \end{subfigure}
  %\vspace{-0.2cm}
  \caption{Impact of the reduction of the width of the sound box on the contour lines. Starting from the original instrument (left), a slice of wood is removed from the main axis (centre) and then the two halves are glued back together (right)}
  \label{fig:Reduction CL}
\end{figure}

\begin{figure}
    \centering
    \begin{subfigure}{0.5\textwidth}
    \centering
    \includegraphics[height=5cm]{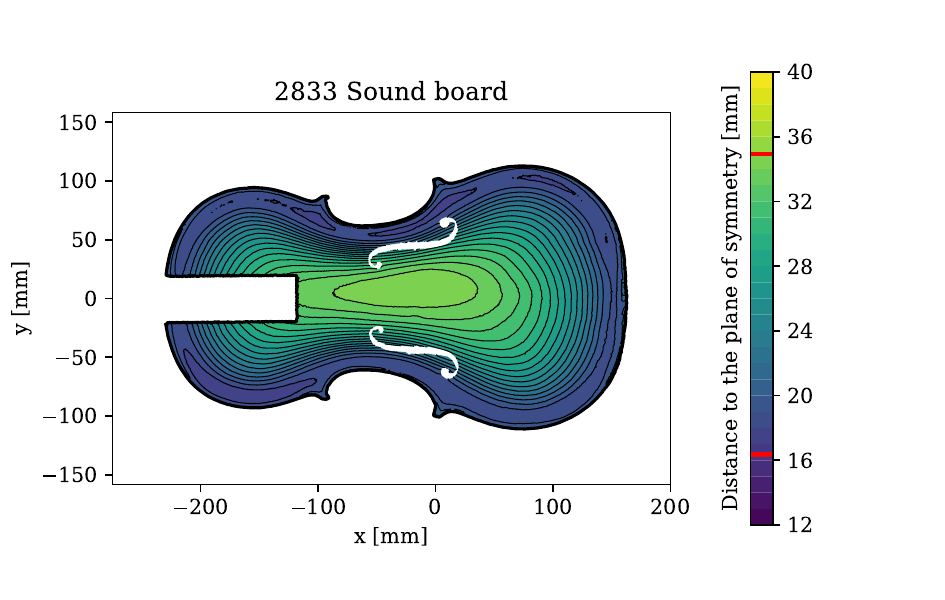}
    \end{subfigure}
    \hfill
    \vspace{-0.75 cm}
    \begin{subfigure}{0.5\textwidth}
    \centering
    \includegraphics[height=5cm]{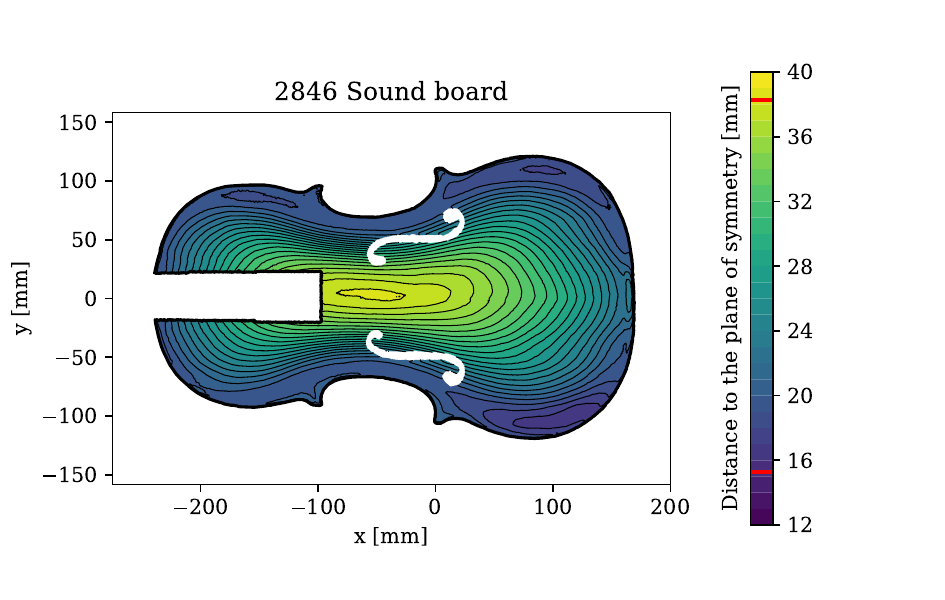}
    \end{subfigure}
    \caption{Contour lines of an unreduced instrument (top) and a reduced one (bottom)}
    \label{fig:Contour Lines}
\end{figure}

\noindent One possible type of reduction consists of removing a wooden slice in the main axis of the instrument and gluing the two halves back together, as described in \cite{romberg1840methode,houssay2015cordes} and shown in Figure \ref{fig:Reduction CL} (from \cite{beghin2024discussion}). This reduction has a direct impact on the contour lines. Whereas they are rather circular for an unreduced instrument (left on Figure \ref{fig:Reduction CL}), they become much sharper and discontinuous at their summit for an instrument reduced at its centre (right on Figure \ref{fig:Reduction CL}). This observation has been verified empirically on violas from the Musical Instruments Museum (MIM, Brussels), as shown in Figure \ref{fig:Contour Lines}. The numbers 2833 and 2846 which appear on the graphs indicate the inventory numbers of the two violas. In a previous work, we have studied a corpus of 38 instruments (25 violins and violas – 13 cellos) on which we have highlighted these differences \cite{beghin2024discussion}. Our core material was photogrammetric meshes, validated and assessed with a submillimetre precision \cite{beghindigital,beghin2023validation}.

\noindent Testimonies about violin reduction are scarce or imprecise. To our knowledge, very few recent studies have been carried out and were almost only qualitative \cite{moens2015voix,houssay2015cordes,echard2022violons}. Luthiers usually examine instruments visually, inspecting both the outer surfaces and the interior (typically using endoscopic cameras) in order to detect traces of alterations and empirically ascertain reductions. This leads us to attempt a more quantitative and objective study of the geometry of these instruments. We also mention the research from \cite{radepont2020revealing} which emphasised traces of reduction on two Andrea Amati's instruments, by computing precisely how much wood had been removed.

\section{Contour lines fitting}

\subsection{Methodology}

As observed qualitatively in our works, the contour lines of an instrument reduced in its centre are less rounded and sharper than the contour lines of an unreduced instrument \cite{beghindigital,beghin2023validation,ceulemans2023baroque,beghin2024discussion}. In this research, we fit contour lines of 25 violins and violas with parametric equations and attempt to observe quantitative differences between (un)reduced instruments. The full list is available in \cite{beghin2024discussion}, except that we removed one dismantled instrument which has no sound board and we added a viola whose photogrammetric mesh was recently acquired. We work exclusively with sound boards, which are the top part of the sound box (as opposed to the back). We are only interested in local contour fitting, hence the use of parabola-like functions. An alternative considered in \cite{piantadosi2017three} is to fit contour lines globally to obtain an equation for the surface of a plate in three dimensional space.%  weighted sums of sines and cosines for violin modelling purposes.

\begin{figure}
    \centering
\includegraphics[scale=0.28]{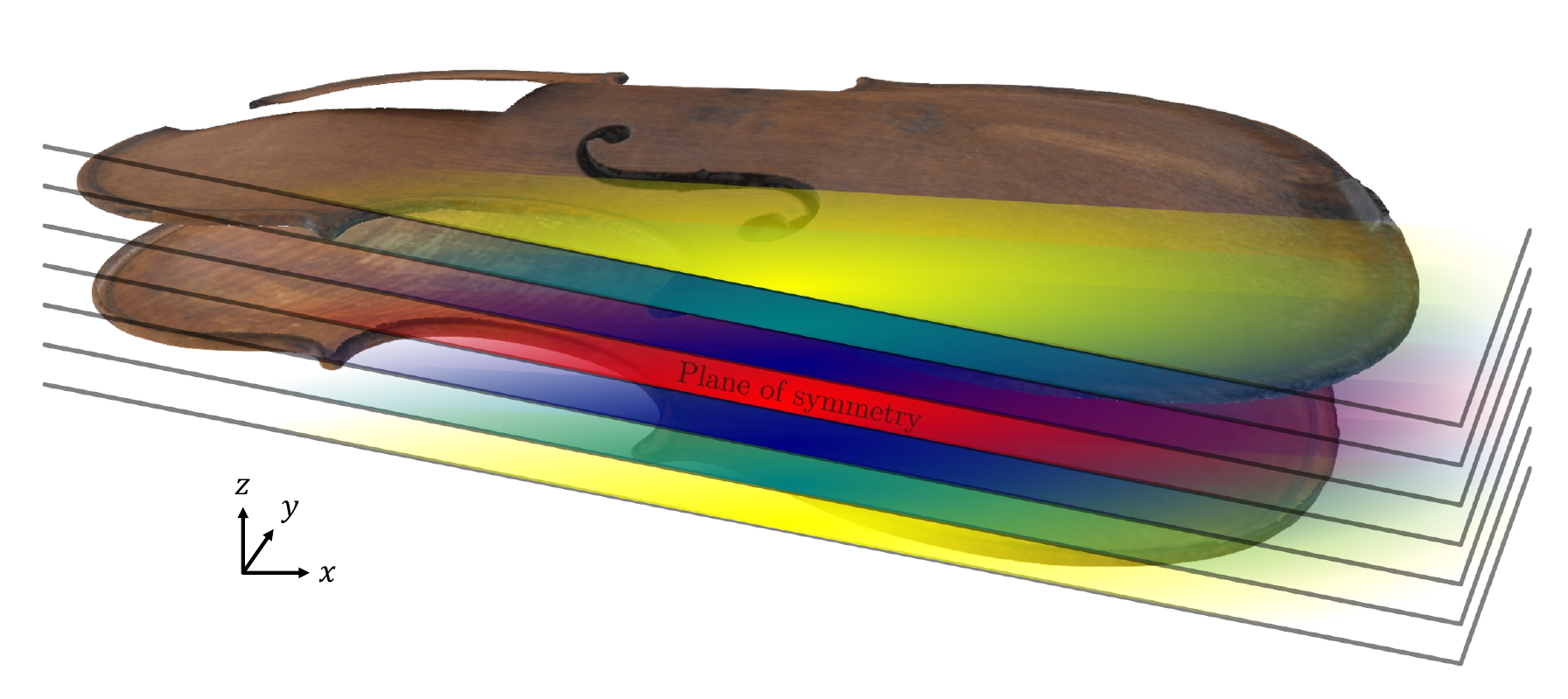}
    \caption{Plane of symmetry between the sound board and the back of a violin and successive parallel horizontal planes}
    \label{fig:g6080}
\end{figure}

\begin{figure}
    \centering
\includegraphics[scale=0.35]{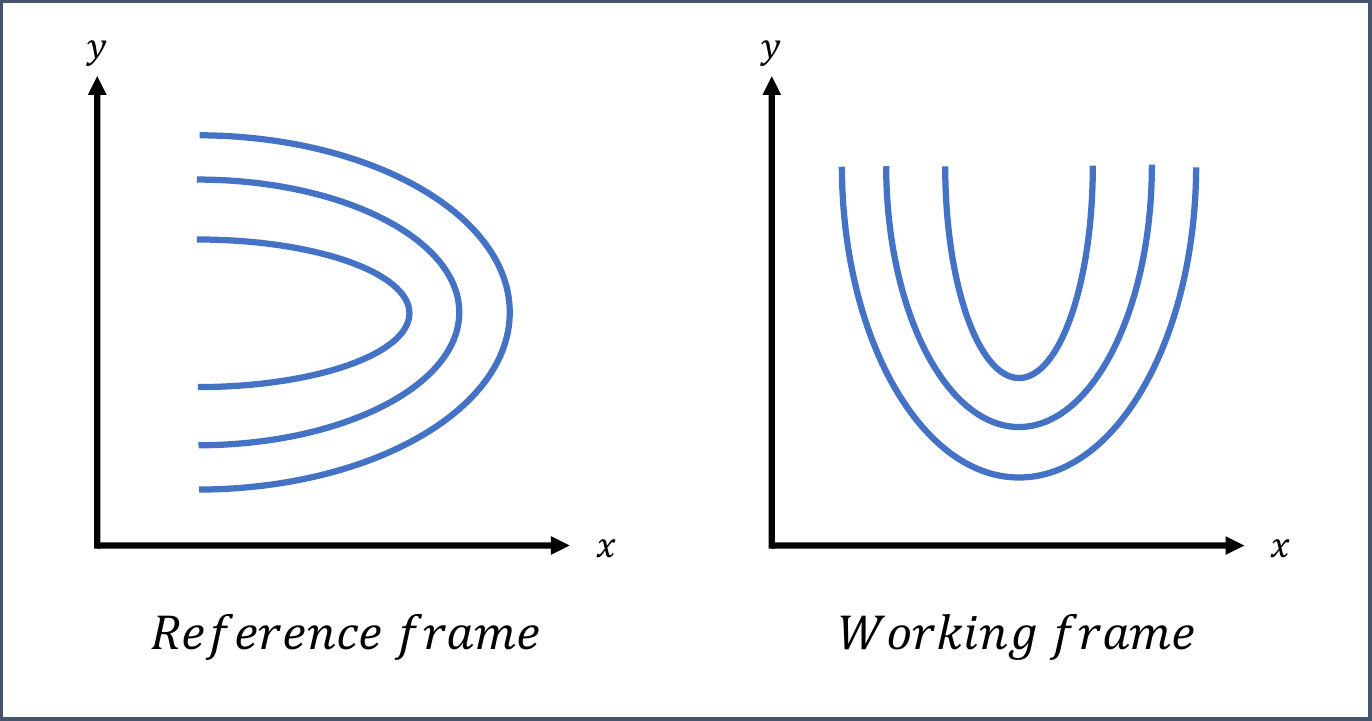}
    \caption{Original reference and new working frame}
    \label{fig:convention frame}
\end{figure}

\begin{figure}
    \centering
\includegraphics[scale=0.45]{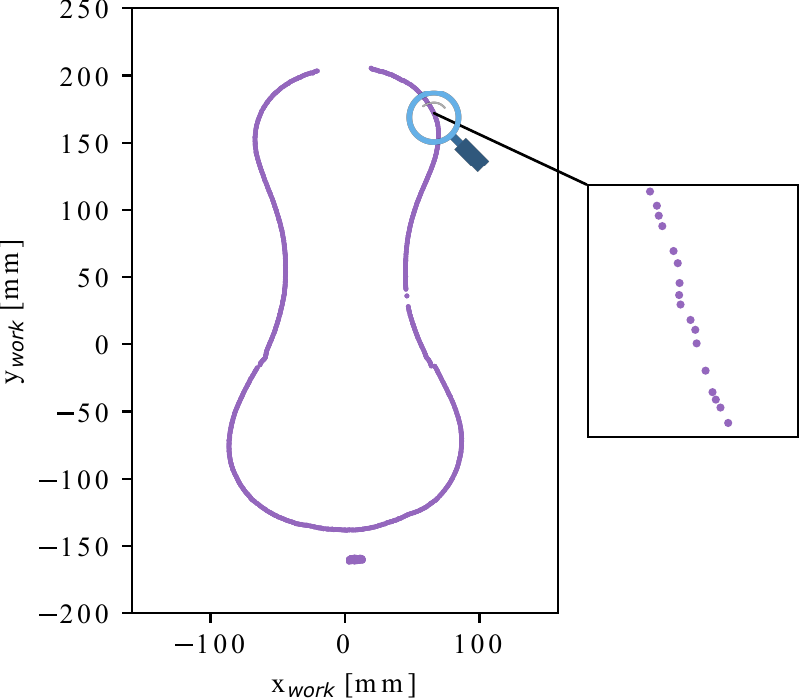}
    \caption{Slice of a non reduced instrument}
    \vspace{-0.7cm}
    \label{fig:Polylines_loupe}
\end{figure}

\begin{figure}
    \centering
    \includegraphics[scale=0.47]{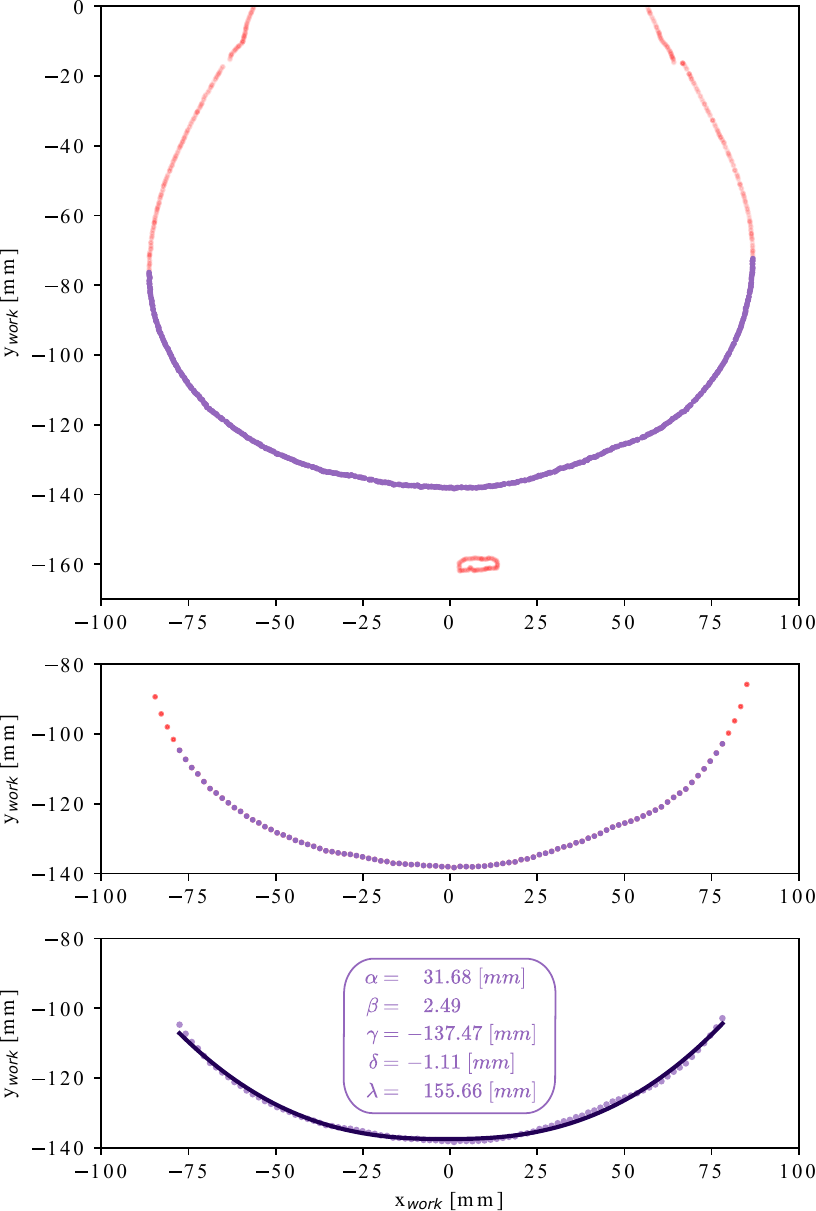}
    \caption{Steps of the pre-processing and fitting of the contour lines}
    \vspace{-0.25cm}
    \label{fig:CL_fit_all}
\end{figure}

\noindent We draw successive horizontal planes intersecting the three-dimensional triangular mesh of the instrument, as shown in Figure \ref{fig:g6080} (from \cite{beghin2024discussion}), with an arbitrarily chosen spacing of \SI{1}{\mm}. The resulting contours are sets of polylines whose vertices belong to edges of the sliced mesh. We then pre-process the polylines to remove the parts that could not be fit (there are artefacts close to the f-holes, for instance), then perform a least-squares fit of the contour according to the parameterised equation of a parabola-like curve. The details of the complete procedure are described hereafter. For the sake of clarity, we rotate the instruments $\ang{90}$ and work in a frame so that the contour lines can be expressed as a function $y = f(x)$ with $x$ pointing to the right and $y$ pointing upwards, as shown in Figure \ref{fig:convention frame} (the rest of our works being based on the frames used in Figures \ref{fig:Contour Lines} and \ref{fig:g6080}). Figure \ref{fig:Polylines_loupe} illustrates a sliced mesh at its $5^{th}$ level, while Figure \ref{fig:CL_fit_all} shows some steps of the pre-processing of the contour lines. Red dots indicate the points that will be deleted in the next step, while the purple ones mark those that will be retained.

\begin{figure}
    \centering
\includegraphics[scale=0.47]{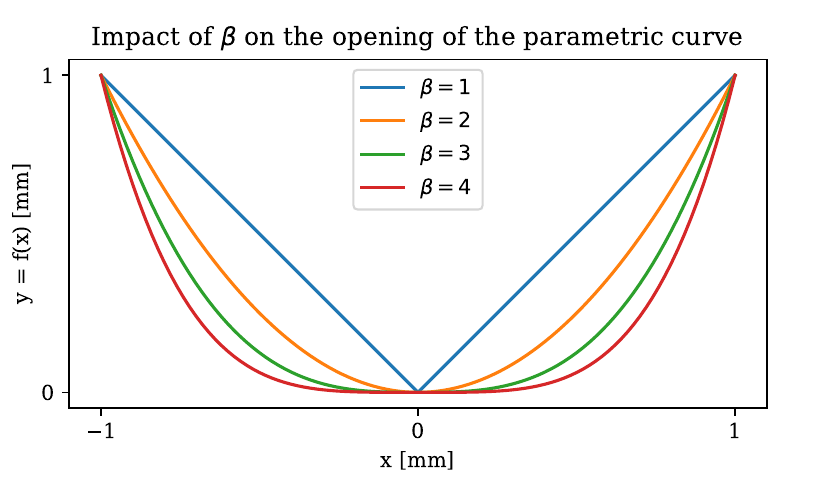}
    \caption{Impact of parameter $\beta$ on the opening of the curve (with $\alpha =1$, $\gamma = \delta = 0$ and $\lambda = 2$)}
    \vspace{-0.7cm}
    \label{fig:several beta}
\end{figure}

\begin{figure}
    \centering
\includegraphics[scale=0.5]{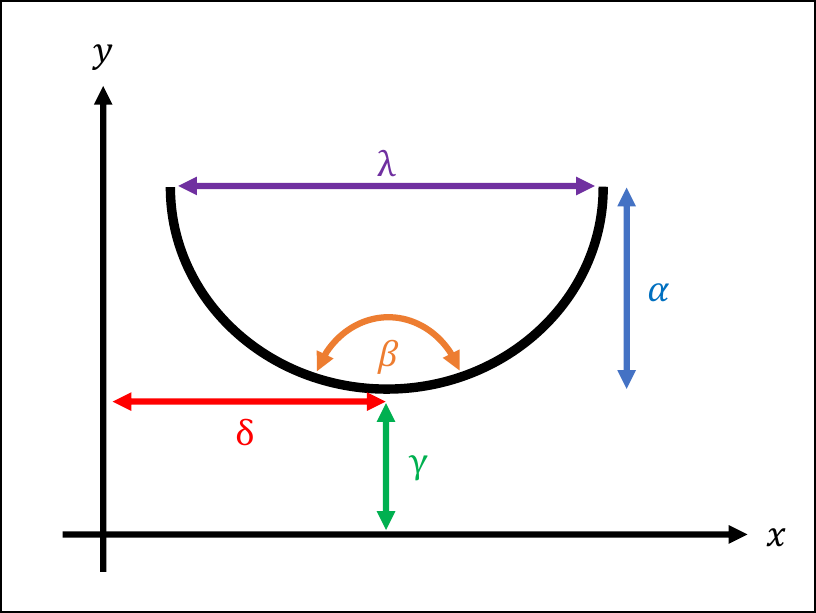}
    \caption{Qualitative impact of the parameters on the $y=f(x)$ curve}
    \vspace{-0.5cm}
    \label{fig:impact parameters}
\end{figure}

\noindent Then the following processing of the polylines is applied:
\begin{itemize}
    \item We keep first the points located at the bottom of the sound board. 
    \item At this stage, the contour lines look like horseshoes (i.e. a 'U' shape with edges that return towards the centre). Since we are interested in fitting the contour lines with a function $y = f(x)$, we can only have one 'y' element at most per 'x'. Hence, we keep all the points between the leftmost and rightmost points (Figure \ref{fig:CL_fit_all}, top, purple points).
    \item In the procedure, some artefacts or discontinuity in the polylines might remain due to the plates which are not perfectly regular surfaces. We automatically detect them and remove them, as shown in Figure \ref{fig:CL_fit_all} (top, red artefacts on the bottom). 
    \item Since the number of points kept on each contour line is different from each other, we resample uniformly each segment with respect to the $x$ axis to keep 100 points. Note that we also tried with 500 and 1000, but the upcoming fitting did not change.
    \item As the relevant part of the contour lines is located at their centre, we have arbitrarily chosen to keep and fit only the points where the angle between the horizontal and the tangent is less than $\ang{60}$. In Figure \ref{fig:CL_fit_all} (centre), we see the 100 regularly spaced points, the ones at the extremities (in red) being removed (tangent greater than $\ang{60}$).
    \item We fit the remaining points to the parameterized function \[ y = f(x) = \alpha \left| \frac{x-\delta}{\frac{\lambda}{2}}\right|^\beta + \gamma \] where $\alpha$ denotes a vertical stretch/compression, $\gamma$ and $\delta$ are vertical and horizontal translations, and exponent $\beta$ is associated with the sharpness of the curve. Figure \ref{fig:several beta} shows the impact of the parameter $\beta$ on the opening and shape of the curve. Roughly speaking, values of $\beta$ smaller than two lead to sharper curves, while larger values give a flatter behaviour. As the contour lines do not have the same width along the sound box, we use a normalisation term $\frac{\lambda}{2}$ in the equation, where $\lambda$ corresponds to the width of the contour. We depict on Figure \ref{fig:impact parameters} how the four fitted parameters $\alpha$, $\beta$, $\gamma$, $\delta$ and the measured width $\lambda$ affect the overall shape of the curve.
    \item The least-squares fit is done using the python function \texttt{scipy.optimize.curve\_fit}. To do so, we initialise $\beta_0 = 2$ (the shape of a parabola) and $\delta_0 = \bar{x}$ as the average of the $x$-component of the points to be fitted. We then set $\tilde{X} = |x-\bar{x}|^2$ and $\tilde{Y} = \gamma_0 $ and initialise $\alpha_0$ and $\gamma_0$ by solving a linear regression on $f(\tilde{X}) = \alpha_0 \tilde{X} + \tilde{Y}$. Figure \ref{fig:CL_fit_all} (bottom) illustrates the results of the fit for a non reduced instrument at its $5^{th}$ level.
    \item We ultimately compute the mean absolute error (MAE) between the points and the predictions and we remove the fits of the whole levels whose MAE is greater than $1.5$. This allows to remove bad fits or levels with artefacts which have not been correctly detected (those artefacts may be caused by wood ageing and twisting or cracking over time).
\end{itemize}

\subsection{Results}

\begin{figure}
    \centering
\includegraphics[scale=0.5]{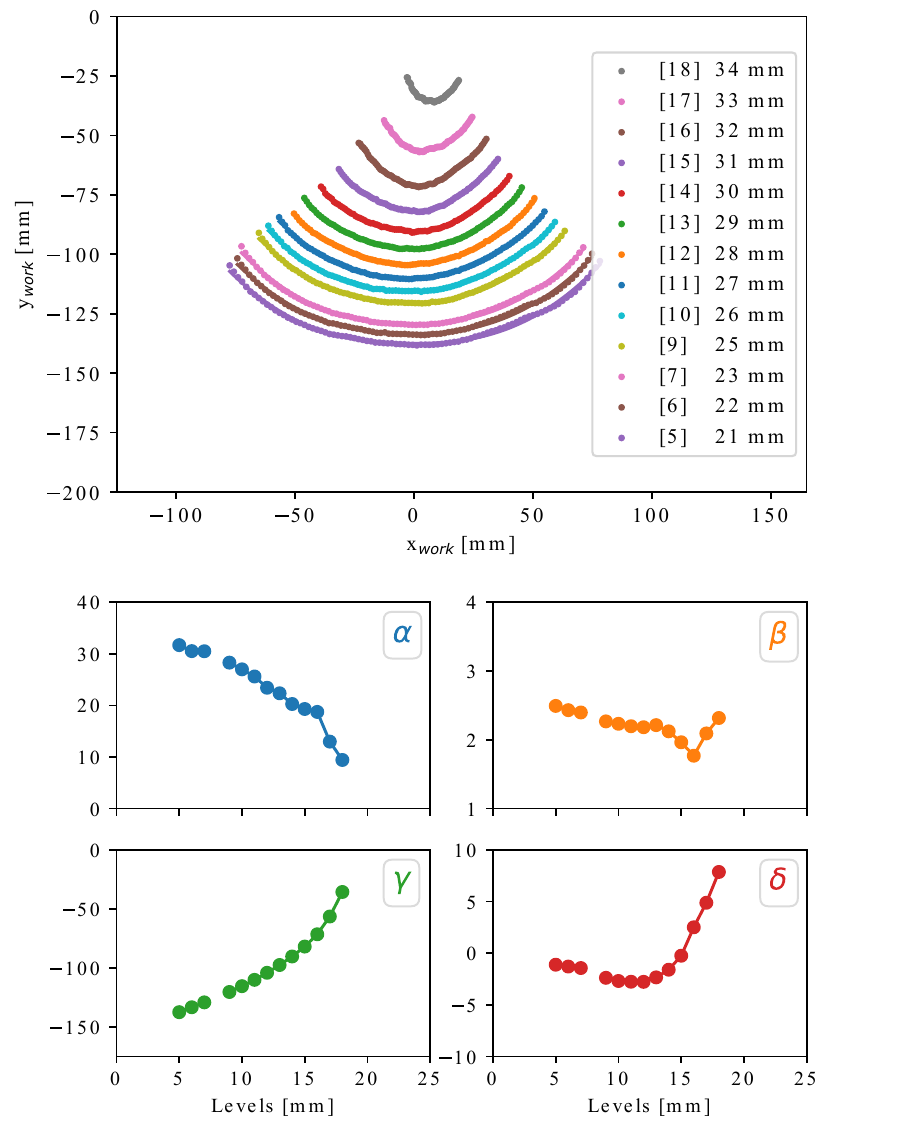}
    \caption{Contour lines fitting of an unreduced violin sound board}
    \label{fig:2833_sb_ID}
\end{figure}

\begin{figure}
    \centering
\includegraphics[scale=0.5]{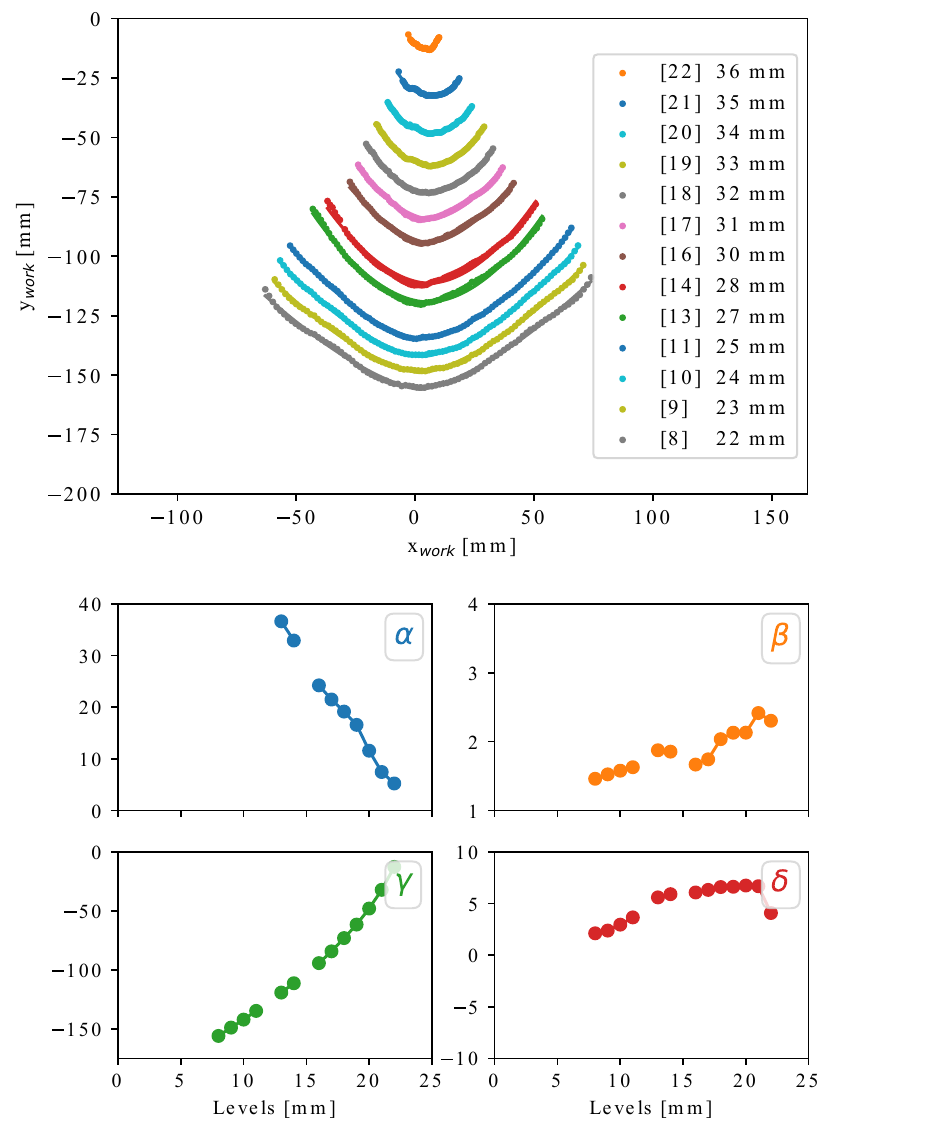}
    \caption{Contour lines fitting of a reduced violin sound board}
    \label{fig:2846_sb_ID}
\end{figure}

Figures \ref{fig:2833_sb_ID} and \ref{fig:2846_sb_ID} illustrate on their top the contour lines of the sound boards of the two instruments already shown in Figure \ref{fig:Contour Lines}. The distance in \SI{}{mm} represents the altitude of the contour line with respect to the plane of symmetry of the violin (an exhaustive discussion about the identification of this plane of symmetry can be found in \cite{beghin2024discussion}). We define the plane of symmetry as the plane that runs through the centre of the instrument’s sound box, parallel to both the sound board (the top plate) and the back (the bottom plate). The number in square brackets indicates the level at which the plate was cut. Level [1] indicates the altitude at which the plate is cut for the first time, starting from the plane of symmetry. Levels and altitudes that do not appear are those for which the mean absolute error (MAE) between the original polyline points and the fit predictions is too high. If we take the example of Figure \ref{fig:2833_sb_ID}, the sound board mesh is first intersected at a height of \SI{17}{mm}, considered as level [1]. However, we need to wait until level [5] (corresponding in this precise case to a height of \SI{21}{mm}) to have the first fit of sufficient quality. This level [5] was actually shown in Figures \ref{fig:Polylines_loupe} and \ref{fig:CL_fit_all}. Figures \ref{fig:2833_sb_ID} and \ref{fig:2846_sb_ID} also display graphically on their bottom the values of the four fitted parameters for each level in four separate boxes (stretch $\alpha$ in blue, opening $\beta$ in orange and translations $\gamma$ and $\delta$ in green and red). Outliers are not represented on those graphs.

\begin{figure}[H]
    \centering
\includegraphics[scale=0.3]{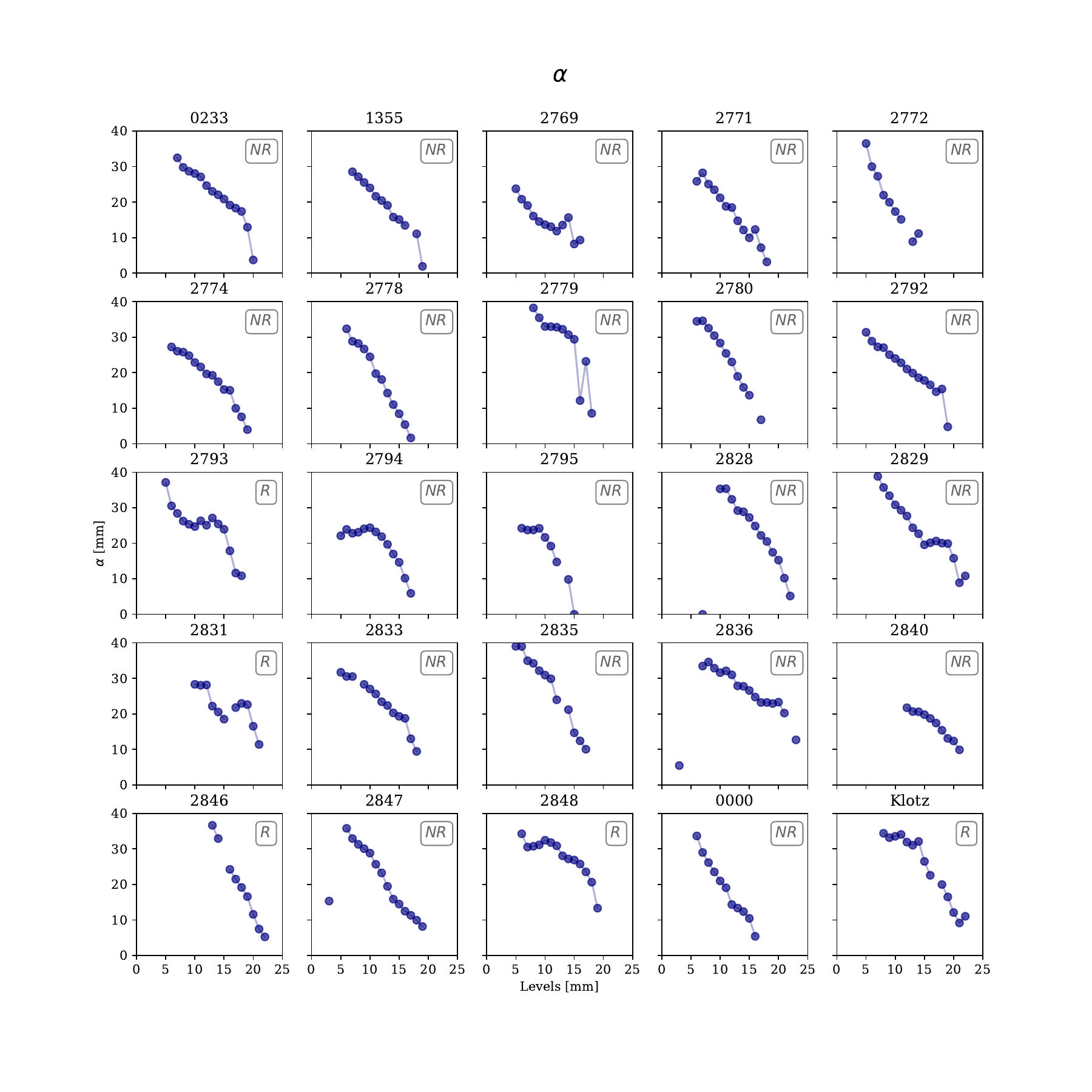}
    \caption{$\pmb{\alpha}$ profiles of the violins and violas of the corpus}
    \label{fig:ALPHA_HAT}
\end{figure}

\begin{figure}[H]
    \centering
\includegraphics[scale=0.3]{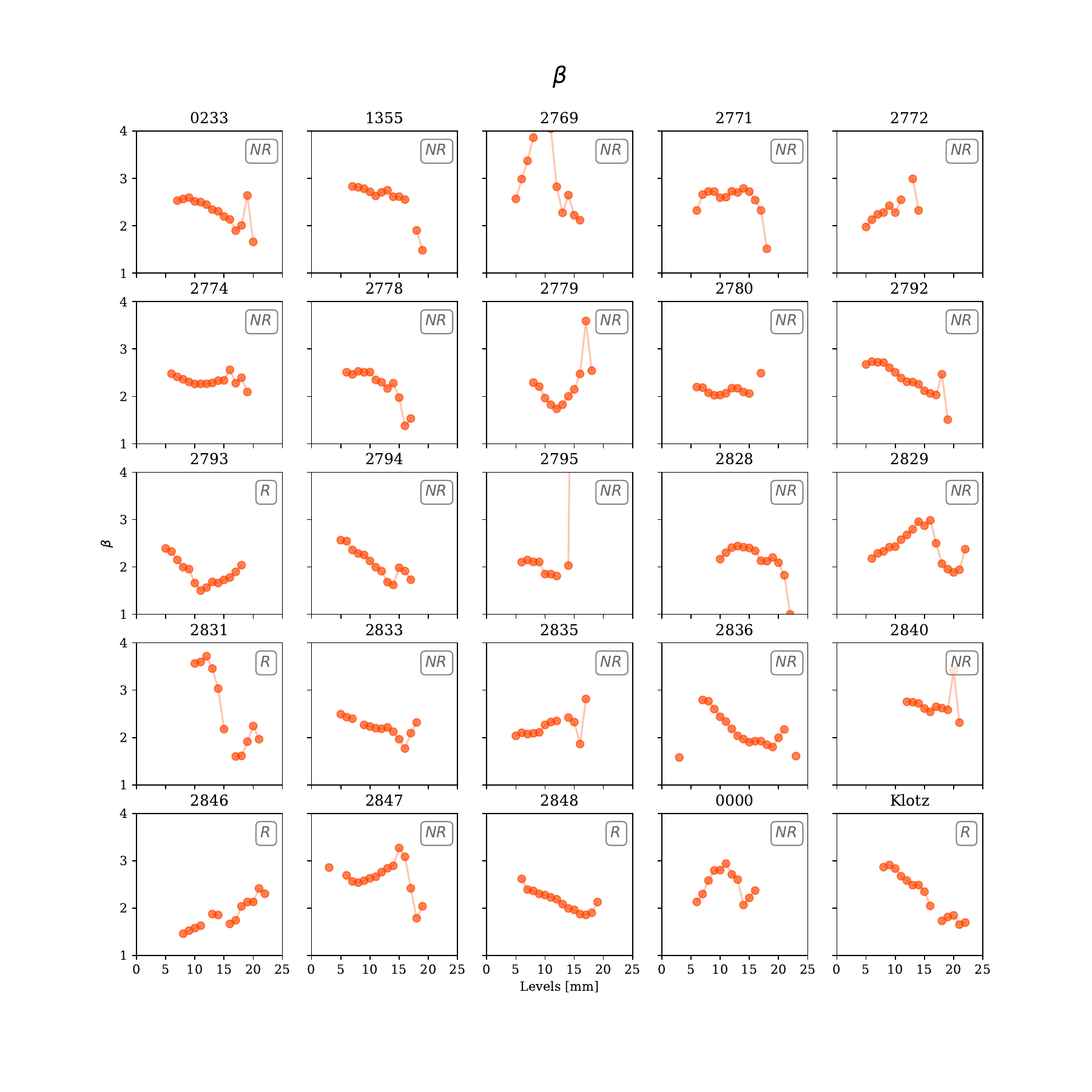}
    \caption{$\pmb{\beta}$ profiles of the violins and violas of the corpus}
    \label{fig:BETA}
\end{figure}

\noindent Figures \ref{fig:ALPHA_HAT} and \ref{fig:BETA} show the whole set of profiles for parameters $\pmb{\alpha}$ and $\pmb{\beta}$ for the 25 sound boards of the violins and violas of our corpus (in the rest of the article, we will use bold notation to designate a vector of parameters). At first sight, we can detect no correlation between $\pmb{\alpha}$ and $\pmb{\beta}$ trends. If we look at the $\pmb{\beta}$ profile of the two instruments we studied above (inventory numbers 2833 and 2846), we see that it tends to increase for the reduced instrument while it decreases for the unreduced instrument. Even if it is somehow complicated to interpret them, we believe that these parameters provide information about whether or not instruments were reduced.

\section{Classification}

\subsection{Model Selection}

\noindent We aim at developing a classification tool to process automatically our data and to observe differences between reduced and unreduced instruments. We want to develop a method that is independent of human opinion and that would objectively and automatically classify those two categories. Our first steps are to choose a model and the features we want to use. Due to the heritage status of our corpus, we are limited to a small number of instruments. More specifically, the sound boards of 25 violins and violas will be analysed (2 photogrammetric meshes of instruments were acquired in 2021, 22 in 2022 and 1 in 2024). The complete acquisition method is described in \cite{beghin2023validation}. 

\noindent Based on interviews with luthiers and musicologists, and under the supervision of Pr. Anne-Emmanuelle Ceulemans, musicologist and curator of European stringed instruments at the Musical Instruments Museum (Brussels), we determined that 20 instruments are most likely not reduced, while the other 5 are (those are denoted NR and R respectively on Figures \ref{fig:ALPHA_HAT} and \ref{fig:BETA}). This analysis was carried out using endoscopic, visual and tactile examinations. As the corpus is too small to make a relevant partition between a training and a test set, we assess the performances of our classifier using a leave-one-out cross validation. This consists in training the model on all instruments except one, and seeing what the model predicts on the one that was held out. The experiment is repeated with each instrument successively held out, to obtain an informed estimate of the classification accuracy of the whole procedure (in particular its generalisation ability). To match the different sizes of the classes (20 unreduced violins vs. 5 reduced) and to avoid bias, we need to balance the two classes in order to achieve a fair evaluation (hyperparameter to be set in the model).

\begin{figure}
    \centering
    \includegraphics[height=5.2cm]{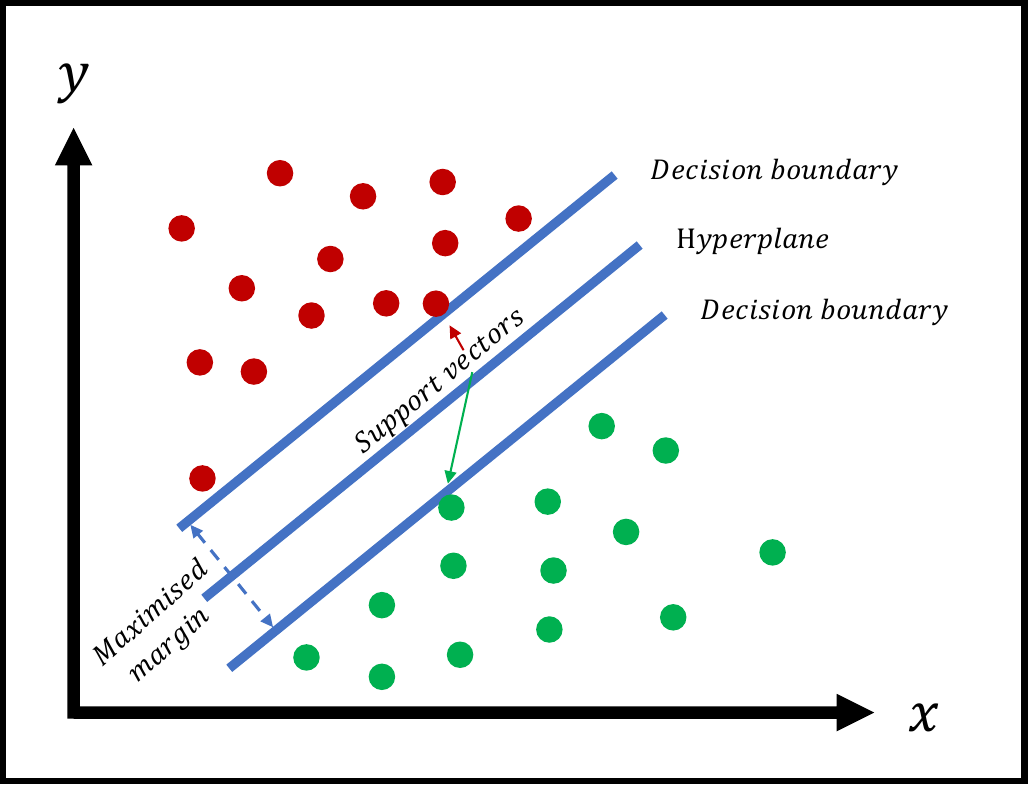}
    \caption{Two-dimensional example of a SVM classifier with a linear kernel}
    \label{fig:SVM example}
\end{figure}

\noindent After some preliminary experiments we decided to focus on a well-known classification method, namely Support Vector Machines (SVM). In a nutshell, a SVM classifier is a supervised learning algorithm whose goal is to find the optimal hyperplane that best separates data points of different classes in a high-dimensional space. Its key concepts, illustrated with a basic two-dimensional example in Figure \ref{fig:SVM example}, are:
\begin{itemize}
    \item \textbf{Margin:} A SVM classifier maximises the margin (i.e. the distance) between data points of different classes and the separating hyperplane.
    \item \textbf{Support Vectors:} These are the data points that lie closest to the decision boundary and are most critical in defining the hyperplane.
    \item \textbf{Linear vs. Non-linear:} If the data is linearly separable, SVM finds a straight line (in 2D) or hyperplane (in higher dimensions). For non-linear data, it uses a kernel to lift data into a higher-dimensional space where it becomes linearly separable. 
    \item \textbf{Regularisation parameter $C$:} The strength of the regularisation is inversely proportional to $C$, a hyperparameter that controls the trade-off between maximizing the margin and minimizing classification error. A small value of $C$ implies a strong regularisation : it prioritises simplicity and generalisation over fitting the training data perfectly, which leads to a wider margin, meaning the model is less sensitive to individual data points (especially outliers). A too strong regularisation might lead to underfitting. Conversely a larger value of $C$ penalises misclassifications more heavily. It will try to fit the training data as well as possible, even if that means a smaller margin, which might increase the risk of overfitting.     
\end{itemize}

\subsection{Features Selection}
\label{Section: features selection}

\subsubsection{Preliminary choices}
\label{Section: preliminary choice}

\begin{figure}[H]
    \centering
\includegraphics[height=4.5 cm]{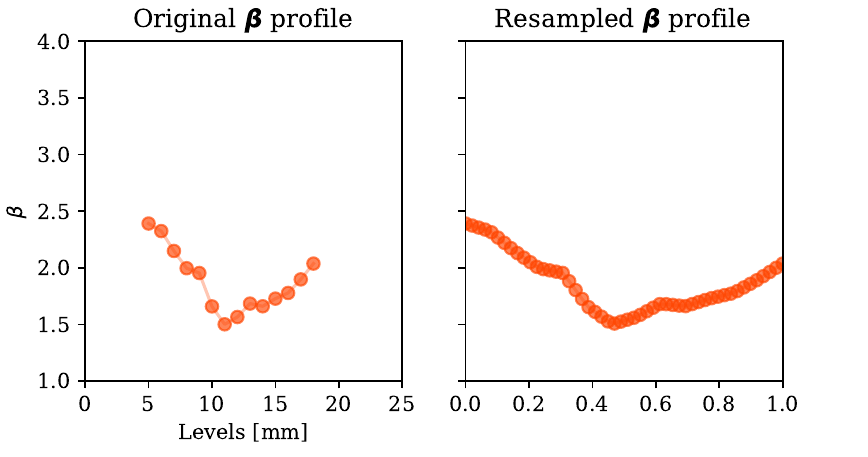}
    \caption{Original $\pmb{\beta}$ profile (left) and resampled one (right)}
    \vspace{-0.5cm}
    \label{fig:BETA_resampling}
\end{figure}

When using a classifier, it is important that the vectors given as input to the model all have the same dimensions. As in this case not all the instruments have the same number of levels (the sound boards have different heights, some levels were rejected because of a bad curve fitting of the contour lines), the data had to be standardised. To do so, we first map the levels to the segment $[0,1]$. Specifically, the first and last “active” levels are mapped to $0$ and $1$ respectively, while the intermediary points are interpolated linearly between those two extremities, based on the positions of vectors $\pmb{\alpha}$, $\pmb{\beta}$, $\pmb{\gamma}$ and $\pmb{\delta}$. We have arbitrarily chosen to sample 50 points equally spaced per vector. In this way, each instrument can be represented through its contour lines as a new vector $\in \mathbb{R}^{200}$ (50 dimensions per parameter). An example of resampling for a $\pmb{\beta}$ profile can be seen in Figure \ref{fig:BETA_resampling}.

\noindent The performance of a SVM model depends on the features given as input, but also on the kernel chosen for that SVM model as well as the regularisation hyperparameter $C$. We have tested different configurations for those, namely
\begin{itemize}
\item A \textit{Linear} kernel or a \textit{Radial Basis Function (RBF)} kernel. Other kernels such as polynomial and sigmoid might also be possible, but the first two are more commonly used.
\item A regularisation hyperparameter $C$ that lies between $C = 10^{4}$ (weak regularisation) and $C = 10^{-4}$ (strong regularisation). 
\item The input features. We have tried to give either the parameters vectors $\pmb{\alpha}$, $\pmb{\beta}$, $\pmb{\gamma}$ and $\pmb{\delta}$ $\in \mathbb{R}^{50}$ individually for each instrument, or all 200 parameters altogether at once.
\end{itemize}

\noindent A final preprocessing step before using our models is to normalise the data. Normalisation in the context of SVM refers to the process of rescaling input features to a common range, typically by centring them around zero and reducing them to unit variance. Normalisation has no effect when we only consider one parameter vectors in $\mathbb{R}^{50}$ at a time. However, it is crucial when we give all 200 parameters at once, or when we add additional derived features. Indeed, SVM models are sensitive to the magnitude of feature values: features with larger numerical ranges can disproportionately influence the decision boundary. For example, before normalisation, values of exponent $\beta$ lie between $1$ and $4$ while values of the translation parameter $\gamma$ lie between $0$ and $150$ (in absolute value). Normalisation of our data will hopefully lead to more stable learning and better generalisation performance.

\begin{figure}
    \centering
    \begin{subfigure}{0.5\textwidth}
    \centering
    \includegraphics[height=5.2cm]{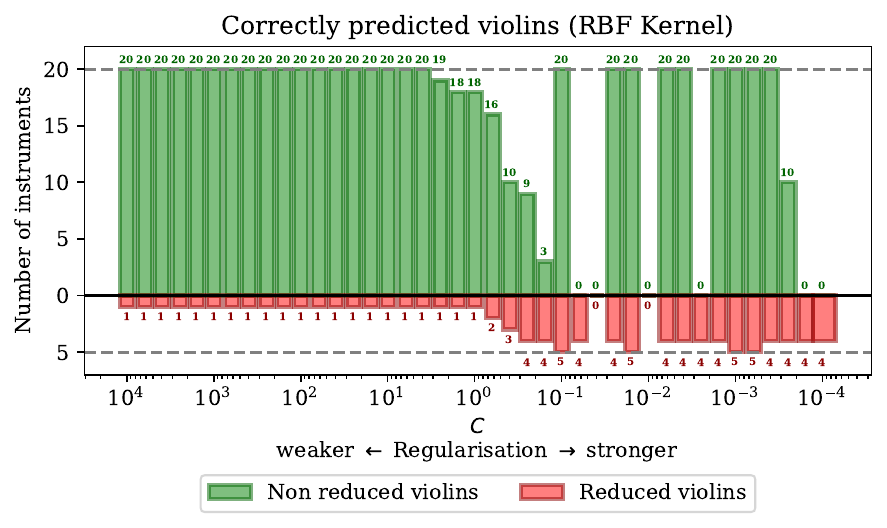}
    \end{subfigure}
    \hfill
    \begin{subfigure}{0.5\textwidth}
    \centering
    \includegraphics[height=5.2cm]{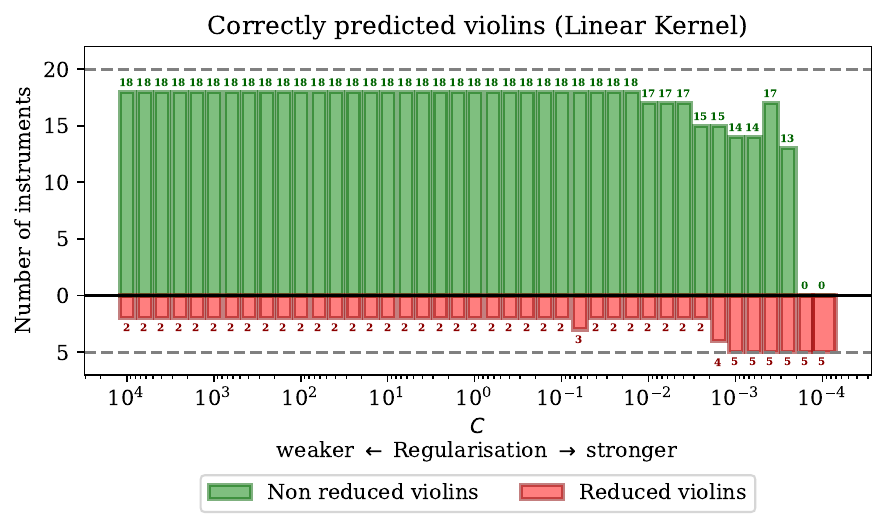}
    \end{subfigure}
    \caption{Performances of two SVM models with RBF (top) and linear (bottom) kernels for different values of the regularisation parameter $C$, with all 200 input features}
    \vspace{-0.5cm}
    \label{fig:200_all}
\end{figure}

\noindent Figure \ref{fig:200_all} shows the classification performance of two SVM models with RBF (top) and linear (bottom) kernels for different values of the regularisation parameter $C$. In green, we see the number of non reduced instruments correctly classified among the 20, and in red, the number of reduced instruments correctly classified among the 5. As a reminder, these results were obtained with a leave-one-out cross validation, i.e. by training the model on 24 instruments, evaluating the performance for the last instrument, and repeating this experiment 25 times (for each value of $C$). Remember that even if the two classes do not have the same number of entities, the red part of the graph is as important as the green part because of the class balance specified as hyperparameter in the SVM model. 

\noindent With a RBF kernel, we can see that the stronger the regularisation ($C$ decreases), the more unstable the model becomes. We see perfect performances (20 + 5 correctly classified instruments) which alternate with very poor ones (0 + 0 correctly classified) for a very small change in $C$. Figure \ref{fig:200_all} (top) shows that $C = 10^{-3}$ seems to be a good choice of regularisation. However, when we take this value of $C$ and use it, for example, to train 23 instruments and predict the last 2, the results are disappointing. With a linear kernel (Figure \ref{fig:200_all}, bottom), the SVM model shows greater stability depending on the values of the regularisation parameter $C$. Its score remains almost constant up to $C = 10^{-2}$, but it never reaches a fully correct performance (20 + 5 instruments correctly classified). We conclude that we will need a better way to automatically find an optimal $C$, as described in Section \ref{Section:23-1-1}. 

\noindent In order to better visualise performance as a function of parameter $C$, we look at the balanced accuracy of the model. Unlike accuracy, a typical measure in machine learning that only considers when a classification model is correct overall, balanced accuracy standardises predictions and takes into account class imbalance. Both definitions are given below, where TP, TN, FP, FN, TPR and TNR stand respectively for the true positive results of the prediction, the true negative, the false positive, the false negative, the true positive rate (number of correct positive predictions among the positive entities) and finally the true negative rate (same with the negative predictions). 
\begin{align*}
    Accuracy = \frac{TP + TN}{TP + TN + FP + FN} &= \frac{0+20}{25} = 80\% \\
    Balanced \ accuracy = \frac{TPR + TNR}{2} &= \frac{\frac{0}{5} + \frac{20}{20}}{2} = 50\%
\end{align*}

\noindent To illustrate we give a numerical example where we classify naively the 25 violins as not reduced. This is objectively a very poor classifier, but still with a relatively high accuracy of 80\%. Its balanced accuracy is only 50\%. 

\begin{figure}
    \centering
    \begin{subfigure}{0.5\textwidth}
    \centering
    \includegraphics[height=5.2cm]{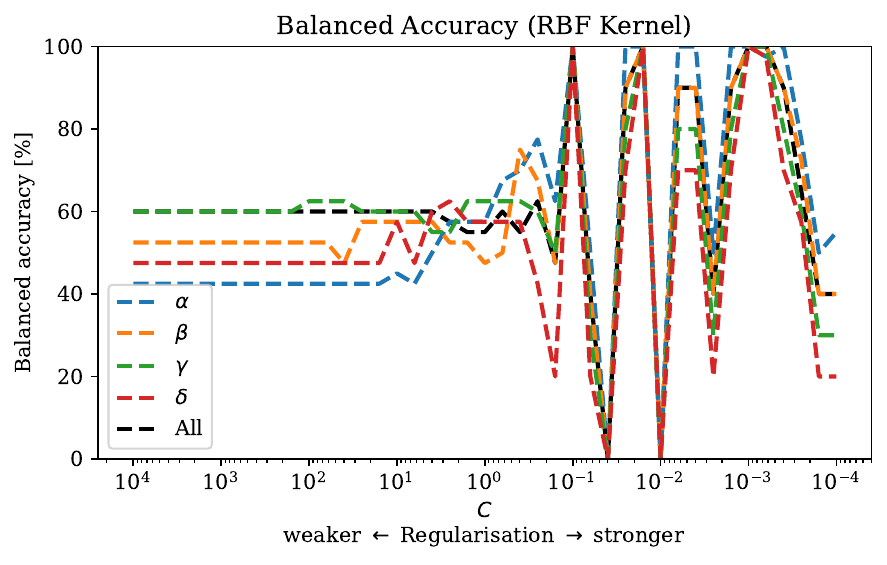}
    \end{subfigure}
    \hfill
    \begin{subfigure}{0.5\textwidth}
    \centering
    \includegraphics[height=5.2cm]{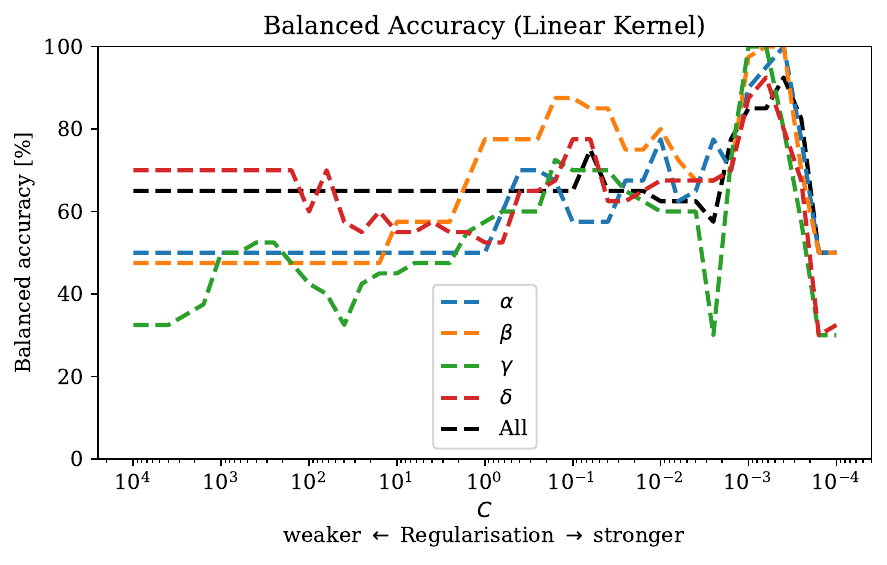}
    \end{subfigure}
    \caption{Balanced accuracy depending on the regularisation parameter $C$ for SVM models with RBF (top) and linear (bottom) kernels}
    \vspace{-0.5cm}
    \label{fig:Balanced_accuracy}
\end{figure}

\noindent Figure \ref{fig:Balanced_accuracy} shows the balanced accuracies for classification with an SVM model and an RBF kernel (top) and a linear kernel (bottom). On both graphs, we see the performance of the classifier (still in leave-one-out cross validation) given the input parameters $\pmb{\alpha}$, $\pmb{\beta}$, $\pmb{\gamma}$ and $\pmb{\delta} \in \mathbb{R}^{50}$, or all four at once. The linear kernel delivers results that are more stable than the RBF, whose performance also oscillates between very good and very poor predictions for the four parameters. We are somehow surprised by the results of that RBF kernel, since it is as difficult to predict 100\% correctly as it is to be wrong every time. Our graphs suggest that it is preferable not to regularise too much with an RBF kernel (to maintain stability and consistency in the method, even though some results might be better with stronger regularisation) and to regularise more for a linear kernel, whose results improve as $C$ decreases. It seems that leave-one-out validation is not the most appropriate method to assess the quality of the predictor, although our small corpus prevents us from proceeding in any other way to reach concrete conclusions. Interestingly, we can see that for a linear kernel with regularisation, parameter $\beta$, which controls the opening of the curve, is more indicative than the others to detect a violin reduced in width, as we would have hoped. Indeed, using balanced accuracy, the performance results only become interesting above 80\%. Below this threshold, the accuracy is quite low (remember that classifying all the entities in the same way is equivalent to a balanced accuracy of 50\%). With an RBF kernel, it is difficult to conclude because no parameter stands out from the others (for a weak regularisation, the accuracy is too low and for a strong regularisation, the method becomes unstable). 

\subsubsection{Features issued from $\beta$ profile}

\begin{figure}
    \centering
\includegraphics[scale=0.33]{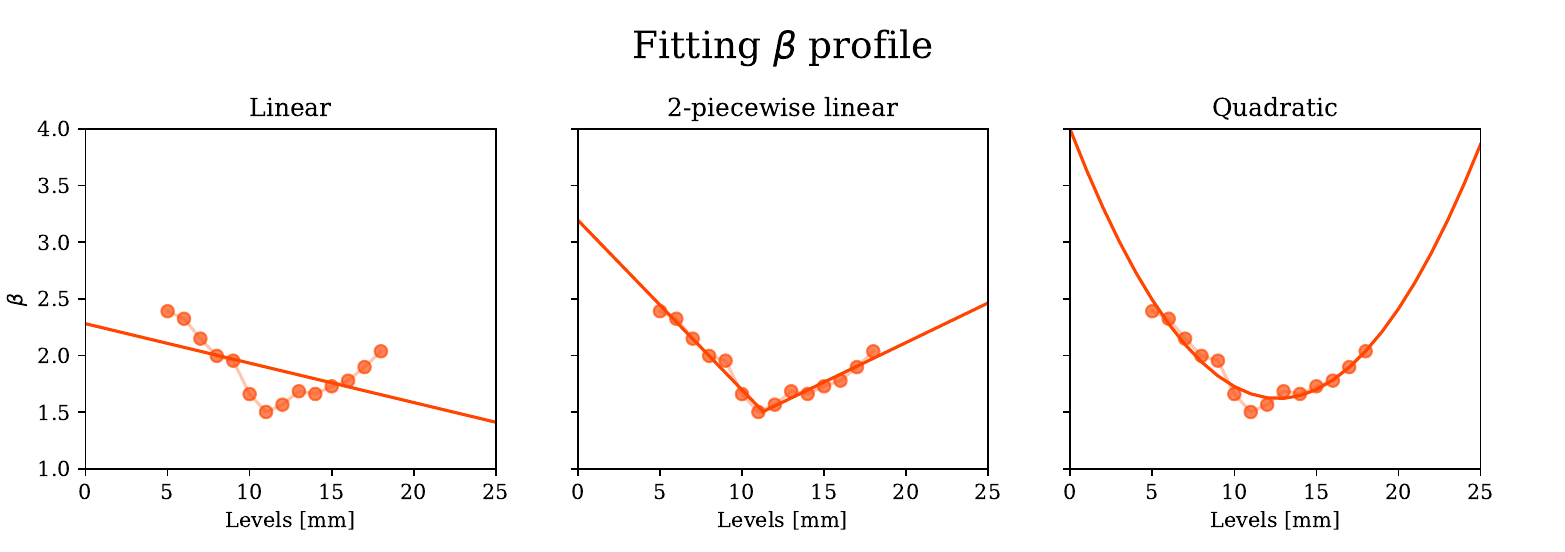}
    \caption{Three fittings of $\pmb{\beta}$ profile. Linear (left), 2-piecewise linear (centre) and quadratic (right).}
    \vspace{-0.5cm}
    \label{fig:Beta_3_fits}
\end{figure}

Preliminary experimental results above showed that the $\beta$ parameter (opening of the parametric curve) is quite significant. To support our study, we also focused on the general characteristics of the complete profile of parameter $\beta$. To derive additional features, we approximated each $\pmb{\beta}$ profile respectively by a straight line (linear fitting), two half straight lines (2-piecewise linear fitting) and by a parabola (quadratic fitting). Those three type fittings are illustrated in Figure \ref{fig:Beta_3_fits}. A linear fitting provides two more features (coefficients of a first order polynomial), a quadratic fitting three (coefficients of a second order polynomial) and the 2-piecewise linear fitting gives five more features : two slopes, two intercepts and one breakpoint, which is the abscissa where the two segments meet. Note that since this procedure ends with the same number of features per instrument, we applied our fittings on the original $\pmb{\beta}$ profile derived from the levels, and not the resampled one. Finally, we also consider the number of parameters $\beta$ above and below a certain threshold within the full profile $\pmb{\beta}$. A standard parabola is defined with $\beta = 2$. We have therefore arbitrarily chosen to count, by instrument and for the full profile, the number of parameters corresponding to $\beta < 2$, to $\beta < 3$ and to $\beta > 3$. We motivate this feature engineering on the $\pmb{\beta}$ profile in order to have a more concise and interpretable classifier. By providing 50 parameters (or even 200) to the model, we do not really know which ones will be decisive, even if a projection into a lower-dimensional space might give an idea on the importance of each feature. With fewer parameters, classification becomes clearer, more concise, and interpretable. Furthermore, it prevents overfitting.

\section{Classification with robust choice of regularisation}
\label{Section:23-1-1}

The prediction results with a leave-one-out cross validation show high instability depending on the choice of the parameter $C$. We first propose in Section \ref{Section: automatic choice of regularisation} a procedure that selects that regularisation parameter $C$ in an automated and more robust way. We will then illustrate the final results of the classifier in Section \ref{Section:results}. A detailed comparison will be established between the features proposed in Section \ref{Section: features selection}.

\subsection{Automatic choice of regularisation}
\label{Section: automatic choice of regularisation}

As we saw in Section \ref{Section: preliminary choice}, the choice of parameter $C$ is a delicate issue that can give rise to underfitting or overfitting problems. A poor choice of $C$ can also lead to strong instability in the SVM method, as is the case with a strong regularisation and an RBF kernel. In order to overcome this problem and avoid arbitrarily selecting the parameter $C$, we propose a method that chooses it automatically, presented in Algorithm \ref{Algo:23-1-1}. This algorithm is still based on standard leave-one-out cross validation, but it includes an additional inner loop whose aim is to optimise the regularisation parameter $C$, whose score is calculated using balanced accuracy in the 24 remaining instruments. In principle we would like to select the value of $C$ producing the best balanced accuracy, serving as a validation score, but Figures \ref{fig:200_all} and \ref{fig:Balanced_accuracy} show that several values of $C$ can be optimal. When several values of parameter $C$ share the best validation score, we have considered 3 ways to select the value of $C$ :
\begin{itemize}
    \item Taking the strongest regularisation (i.e. the smallest $C$),
    \item Taking the weakest regularisation (i.e. the highest $C$), 
    \item Taking a random $C$. In order to prevent randomness bias with this third way, we repeated the experiments 25 times. 
\end{itemize}

\begin{algorithm*}[t]
\DontPrintSemicolon
\KwData{25 instruments}
\KwResult{Leave-one-out evaluation on each instrument and balanced accuracy}
\ForEach{instrument \circled{T} out of the 25}{
    \ForEach{$C \in \{10^{-4}, \dots, 10^4\}$}{
        \ForEach{instrument \circled{V} from the 24 others (excluding \circled{T})}{
        \begin{itemize}
            \item Train the model on the 23 other instruments (excluding \circled{V} and \circled{T})
            \item Validate the model i.e. check whether it is correct on \circled{V} 
        \end{itemize}
        }
        Define validation score for this value of $C$ as  balanced accuracy over all \circled{V}s
    }
    \begin{itemize}
        \item Select the value of parameter $C$ with the best validation score
        \item Train a new model with this value of $C$ on the 24 instruments (excluding \circled{T})
        \item Validate final model i.e. check whether it is correct on the instrument \circled{T}
        \end{itemize}
}
Compute final overall balanced accuracy on 25 instruments \circled{T}
\caption{Robust regularisation choice with per instrument cross-validation}
\label{Algo:23-1-1}

\end{algorithm*}   

\subsection{Results}
\label{Section:results}

Using Algorithm \ref{Algo:23-1-1}, we evaluate the performances of several subsets of features described in Section \ref{Section: features selection}. More specifically, we draw up a comparative table to establish the classification score (evaluated in terms of balanced accuracy) of our models with an increasing number of features given as input. We combine some sets of features (for example, the two features of a linear fit with the three features of the numbers of values of $\beta$ satisfying $\beta \leq 2$, $\beta \leq 3$ and $\beta > 3$) to see if this has an impact on the scores. The number and type of "raw" features (cfr. Section \ref{Section: features selection}) used for the combinations are : 
\begin{itemize}
    \item 2 - linear fitting,
    \item 3 - quadratic fitting,
    \item 3 - number of values of $\beta$ satisfying $\beta \leq 2$, $\beta \leq 3$ and $\beta > 3$ per $\pmb{\beta}$ profile,
    \item 3 - proportion of values of $\beta$ satisfying $\beta \leq 2$, $\beta \leq 3$ and $\beta > 3$ per $\pmb{\beta}$ profile,
    \item 5 - piecewise linear fitting (2 slopes, 2 intercepts, 1 breakpoint), 
    \item 50 - resampled profiles of $\pmb{\alpha}$, $\pmb{\beta}$, $\pmb{\gamma}$ and $\pmb{\delta}$.
\end{itemize}

\noindent Preliminary results obtained earlier on a limited set of three reduced instruments indicated that the second slope of the piecewise linear fitting appeared to be a reliable indicator, in combination with the number (or proportion) of $\beta$ above/below a certain threshold. We have therefore included three additional combinations: that second slope along with the number (or proportion) of cases where $\beta \leq 2$; the previous two variables plus the number (or proportion) of cases where $\beta \leq 3$; and all three variables together with the number (or proportion) of cases where $\beta > 3$. Note that all these features have been normalised as explained in Section \ref{Section: preliminary choice}, with two exceptions: proportions are ratios between 0 and 1, hence are already scaled, and similarly the breakpoint abscissa from the 2-piecewise linear fitting was mapped onto a segment [0,1] whose endpoints are the first and last active levels of the $\pmb{\beta}$ profile, as shown in Figure \ref{fig:BETA_resampling}.  %It therefore did not need to be renormalised. 
All balanced accuracies for those combinations of features can be found in Table \ref{tab:scores_23_1_1} (those greater than 90\% are highlighted in bold).

\begin{table*}
\centering
\resizebox{\textwidth}{!}{
\begin{tabular}{|l|cc|cc|cc|}
\hline
& \multicolumn{6}{c|}{\textbf{Balanced accuracy} (in $\%$)} \\
\textbf{Number and type of features} & \multicolumn{2}{c|}{\textbf{Random $C$}} & \multicolumn{2}{c|}{\textbf{max $C$ (weak reg.)}} & \multicolumn{2}{c|}{\textbf{min $C$ (strong reg.)}} \\
            & RBF & Lin. & RBF & Lin. & RBF & Lin. \\
\hline
2 (linear fit.) & 41.3 & \textbf{91.3} & 0 & \textbf{100} & 50 & \textbf{100} \\
2 (second slope + number of $\beta \leq 2$) & 75.4 & 75.2 & 50 & \textbf{90} & \textbf{100} & \textbf{90} \\
2 (second slope + proportion of $\beta \leq 2$)  & 73.9 & 77.7 & 50 & 75 & \textbf{100} & \textbf{100} \\
3 (quadratic fit.) & 68.2 & 60.5 & 40 & 50 & \textbf{90} & 70 \\
3 (number of $\beta \leq 2$, $\beta \leq 3$ and $\beta > 3$) & 59.4 & 68.2 & 40 & \textbf{90} & 42.5 & 42.5 \\
3 (proportion of $\beta \leq 2$, $\beta \leq 3$ and $\beta > 3$) & 49.6 & 65.5 & 20 & 40 & 70 & \textbf{90} \\
3 (second slope + number of $\beta \leq 2$ and $\beta \leq 3$) & 69.9 & 63.4 & 40 & 70 & \textbf{90} & 70 \\
3 (second slope + proportion of $\beta \leq 2$ and $\beta \leq 3$) & 75.1 & 80.9 & 50 & 75 & \textbf{100} & \textbf{100} \\
4 (second slope + number of  $\beta \leq 2$, $\beta \leq 3$ and $\beta > 3$) & 62.1 & 60.3 & 40 & 70 & 62.5 & 32.5 \\
4 (second slope + proportion of  $\beta \leq 2$, $\beta \leq 3$ and $\beta > 3$) & 77 & 81.7 & 50 & 75 & \textbf{100} & \textbf{100} \\
5 (linear fit. + number of $\beta \leq 2$, $\beta \leq 3$ and $\beta > 3$) & 63.8 & 69.9 & 30 & \textbf{90} & 77.5 & 52.5 \\
5 (linear fit. + proportion of $\beta \leq 2$, $\beta \leq 3$ and $\beta > 3$) & 39.6 & \textbf{94.6} & 0 & \textbf{100} & 50 & \textbf{100} \\
6 (quadratic fit. + number of $\beta \leq 2$, $\beta \leq 3$ and $\beta > 3$) & 72.2 & 63.4 & 40 & 72.5 & \textbf{90} & 45 \\
6 (quadratic fit. + proportion of $\beta \leq 2$, $\beta \leq 3$ and $\beta > 3$) & 67.8 & 69.6 & 40 & 62.5 & \textbf{90} & 80 \\
5 (piecewise linear fit.) & 80.2 & \textbf{91.2} & 50 & \textbf{100} & \textbf{100} & \textbf{100} \\
8 (piecewise linear fit. + number of $\beta \leq 2$, $\beta \leq 3$ and $\beta > 3$) & 60.2 & 62.2 & 40 & 67.5 & 45 & 45 \\
8 (piecewise linear fit. + proportion of $\beta \leq 2$, $\beta \leq 3$ and $\beta > 3$) & 76.8 & \textbf{95} & 50 & \textbf{100} & \textbf{100} & \textbf{100} \\
8 (piecewise lin. + quadratic fit.) & 73.3 & \textbf{94.6} & 40 & \textbf{100} & \textbf{90} & \textbf{100} \\
13 (lin. fit. + piecewise lin. fit. + quad. fit. + number of $\beta$) & 64.4 & 71.8 & 40 & 80 & \textbf{90} & 70 \\
13 (lin. fit. + piecewise lin. fit. + quad. fit. + proportion of $\beta$) & 64.4 & 82.5 & 30 & \textbf{90} & 80 & \textbf{90} \\
\hdashline
50 (resampled $\pmb{\alpha}$ profile) & 75.8 & \textbf{100} & 50 & \textbf{100} & \textbf{100} & \textbf{100} \\
50 (resampled $\pmb{\beta}$ profile) & 74.6 & \textbf{100} & 40 & \textbf{100} & \textbf{90} & \textbf{100} \\
50 (resampled $\pmb{\gamma}$ profile) & 58.6 & 53.4 & 30 & 65 & 80 & 44.5 \\
50 (resampled $\pmb{\delta}$ profile) & 57.9 & 80 & 20 & 80 & 70 & 80 \\
200 (resampled $\pmb{\alpha}$, $\pmb{\beta}$, $\pmb{\gamma}$ and $\pmb{\delta}$ profiles) & 72.6 & \textbf{92.5} & 40 & \textbf{92.5} & \textbf{90} & \textbf{92.5} \\
\hline
\end{tabular}
} % fin du resizebox
\caption{Balanced accuracies of SVM models with an RBF or a linear kernel trained on different subsets of features (accuracies greater or equal to $90\%$ in bold)}
\vspace{-0.5cm}
\label{tab:scores_23_1_1}
\end{table*}

\noindent First of all, we can see that we achieve remarkable classification performance (for some combinations of features, several balanced accuracies even reach 100\%). This performance is highly dependent on regularisation, but we use a robust algorithm so that we do not have to worry about the choice of $C$, which is computed automatically. For a linear kernel, both weak and strong regularisations seem to perform well, while only the strong regularisation seems to be required to obtain good RBF results. We remain cautious, however, as we have seen that the RBF kernel leads to instabilities and potentially erroneous results. For example, with the first leave-one-out cross validation used in Section \ref{Section: preliminary choice}, training 22 or 23 instruments out of the 24 trained previously with $C=10^{-3}$ (balanced accuracy of 100\% for all resampled parameters in Figures \ref{fig:200_all} and \ref{fig:Balanced_accuracy}) was not at all effective.

\noindent Although the $\pmb{\beta}$ profile was already promising and slightly better than the other three for the initial leave-one-out validation, we see that with our robust algorithm, it reaches now almost perfect classification for a model with a linear kernel and it is still very high for an RBF kernel, especially with strong regularisation (see the penultimate four lines of Table \ref{tab:scores_23_1_1}). Surprisingly, the $\pmb{\alpha}$ profile also gives excellent results. However, it might not be necessary to provide all this information (50 values) to our models. Indeed, smaller number of features derived from the $\pmb{\beta}$ profiles (or their combinations) can also provide very good results. The piecewise linear fitting seems to be a very informative feature, also when combined either with a quadratic fitting or with the proportion of values of $\beta$ below/above a threshold. We wonder whether a small number of features derived from the $\pmb{\alpha}$ profile, similar to those that were computed from the $\pmb{\beta}$ profile, could also give excellent results, a question we leave for further research.

\noindent Finally, we note that a random choice of $C$ does not give significantly better results than weak or strong regularisation for linear kernels or strong regularisation for RBF kernels. Nevertheless, it achieves very high performance (with a linear kernel) for several features, and this performance is sometimes extremely close, or even equal, to that achieved with strong or weak regularisation. We report perfect classification with resampled $\pmb{\alpha}$ and $\pmb{\beta}$ profiles, and very good results with combinations of linear fitting, proportions of values of $\beta$ above/below a threshold and 2-piecewise linear fitting.

\noindent Based on our results, we recommend a linear kernel over an RBF kernel. This preference is primarily due to the linear kernel's greater interpretability, as classification is performed using a hyperplane that separates the data. Additionally, it demonstrates greater stability compared to the RBF kernel. A weak regularisation —corresponding to smaller margins in the SVM classifier— also appears preferable to stronger regularisation, which tends to produce excessively wide margins and leads to overgeneralisation, often classifying all data into the same category, especially for the RBF kernel. However, it is worth noting that despite its advantages, the linear kernel requires significantly more computation time when used with our robust algorithm. Processing a single set of features takes approximately one hour, mainly due to the large number of hyperplanes that must be evaluated (25 × 24 × 23 × the number of C values tested). In contrast, even with an equivalent amount of computation, the RBF kernel appears to be much faster, requiring only about one minute per feature set.

\section{Conclusion}

Our aim was to develop a classification tool based on contour lines features to automatically detect and exploit observed differences between reduced and unreduced violins. This was not straightforward because the sets of characteristics of the instruments are not uniform (not the same number of levels, all levels might not necessarily have a correct fit, etc.). Furthermore, the contour lines in our corpus do not all have "U" or "V" shapes as pronounced as in Figures \ref{fig:Contour Lines}, \ref{fig:2833_sb_ID} and \ref{fig:2846_sb_ID}. A whole spectrum of more or less transformed violins exists for which it is more difficult to quantify the reduction. Some have not been reduced, but simply show deformations caused by ageing (wood warps over time, some instruments have cracked, etc.).

\noindent This approach to detection of instrument reduction is, to the best of our knowledge, completely new and we tried to identify the best features to provide to our models. It seems that the parameter $\beta$, which governs the opening of the fitted contour curve as well as the features derived from its profile are very indicative. In addition, those low-dimensional sets of parameters potentially allow us interpret classifiers which achieve very high performance with our robust algorithm. In a way, we are confident in our tool after selection of the best regularisation parameter.

\noindent In future work, we would like to extend our analysis to violin backs (the lower plate of the sound box), and not just sound boards. This would offer two additional profiles of contour lines per instrument to evaluate (for sound boards, the top is not accessible due to the fingerboard which is removed on the meshes). We have also acquired photogrammetric meshes of cellos, and it would be useful to mix their features with violins and violas to reinforce our conclusions and our digital classification tool. Additionally, we could also test other classification models. Support Vector Machines were chosen because they are a standard tool whose effectiveness has been repeatedly documented in the literature. However, other types of basic classifiers exist, such as decision trees, or more architecturally complex ones such as neural networks, that could be investigated. Finally, additional features could be investigated to enhance classification, including geometric descriptors such as the channel of minima on the plates (see \cite{beghindigital,beghin2023validation,ceulemans2023baroque,beghin2024discussion}), as well as non-geometric characteristics such as surface texture data obtained through photogrammetry.

%-------------------------------------------------------------------------
% bibtex
\bibliographystyle{eg-alpha-doi} 
\bibliography{egbibsample}       

\newcommand{\etalchar}[1]{$^{#1}$}
\begin{thebibliography}{\uppercase{BCFG23}}

\bibitem[BCFG23]{beghin2023validation}
\textsc{Beghin P., Ceulemans A.-E., Fisette P., Glineur F.}:
\newblock Validation of a photogrammetric approach for the objective study of early bowed instruments.
\newblock \emph{Heritage Science 11}, 1 (2023), 170.

\bibitem[BCG24]{beghin2024discussion}
\textsc{Beghin P., Ceulemans A.-E., Glineur F.}:
\newblock A discussion about violin reduction: geometric analysis of contour lines and channel of minima.
\newblock \emph{arXiv preprint arXiv:2404.01995} (2024).

\bibitem[Beg21]{beghindigital}
\textsc{Beghin P.}:
\newblock \emph{A digital tool at the service of organology: validation of a photogrammetric approach}.
\newblock Master's thesis, Ecole polytechnique de Louvain, UCLouvain, 2021.

\bibitem[CBF{\etalchar{*}}23]{ceulemans2023baroque}
\textsc{Ceulemans A.-E., Beghin P., Fisette P., Glineur F., Thys I.}:
\newblock Baroque violas with reduced soundboxes: An evaluation method.
\newblock \emph{The Galpin Society Journal 76 (LXXVI)} (2023), 109--126.

\bibitem[{\'E}ch22]{echard2022violons}
\textsc{{\'E}chard J.-P.}:
\newblock Les violons de {C}r{\'e}mone {\`a} la cour des derniers valois.
\newblock \emph{La Cour en f{\^e}te dans l'Europe des Valois} (2022), 221--235.

\bibitem[Her03]{herzog2003quinton}
\textsc{Herzog M.}:
\newblock \emph{The quinton and other viols with violin traits (16-18th centuries)}.
\newblock PhD thesis, Bar-Ilan University, 2003.

\bibitem[Hou15]{houssay2015cordes}
\textsc{Houssay A.}:
\newblock Cordes fil{\'e}es et violons en italie au {XVII}e si{\`e}cle: quelques cas d’instruments cr{\'e}monais recoup{\'e}s.
\newblock \emph{Duron, Jean; G{\'e}treau, Florence. L'orchestre {\`a} cordes sous Louis XIV : Instruments, r{\'e}pertoires, singularit{\'e}s.} (2015), 139--162.

\bibitem[Moe15]{moens2015voix}
\textsc{Moens K.}:
\newblock Les voix m{\'e}dianes dans l'orchestre fran{\c{c}}ais sous le r{\`e}gne de {L}ouis {XIV} : les instruments conserv{\'e}s comme source d'information.
\newblock \emph{Duron, Jean; G{\'e}treau, Florence. L'orchestre {\`a} cordes sous Louis XIV : Instruments, r{\'e}pertoires, singularit{\'e}s.} (2015), 119--138.

\bibitem[Pia17]{piantadosi2017three}
\textsc{Piantadosi S.}:
\newblock Three dimensional mathematical modeling of violin plate surfaces: An approach based on an ensemble of contour lines.
\newblock \emph{{PLoS} {O}ne 12}, 2 (2017), e0171167.

\bibitem[R{\'E}O{\etalchar{*}}20]{radepont2020revealing}
\textsc{Radepont M., {\'E}chard J.-P., Ockerm{\"u}ller M., de~la Codre H., Belhadj O.}:
\newblock Revealing lost 16th-century royal emblems on two {A}ndrea {A}mati’s violins using {XRF} scanning.
\newblock \emph{Heritage Science 8}, 1 (2020), 1--12.

\bibitem[Rom40]{romberg1840methode}
\textsc{Romberg B.}:
\newblock \emph{M{\'e}thode de violoncelle, adopt{\'e}e par le directeur du {C}onservatoire royal de {P}aris, {\`a} l’usage des classes de cet {\'e}tablissement}.
\newblock H. Lemoine, Paris, 1840.

\bibitem[Sib85]{sibire1885chelonomie}
\textsc{Sibire S.-A.}:
\newblock \emph{La Ch{\'e}lonomie, ou, le parfait luthier}.
\newblock A. Loosfelt, 1885.

\bibitem[Sto90]{stowell1990violin}
\textsc{Stowell R.}:
\newblock \emph{Violin technique and performance practice in the late eighteenth and early nineteenth centuries}.
\newblock Cambridge University Press, 1990.

\bibitem[Zel19]{zeller2019reconstructing}
\textsc{Zeller M.}:
\newblock Reconstructing lost instruments: Praetorius’s syntagma musicum and the violin family c. 1619.
\newblock \emph{De musica disserenda 15}, 1-2 (2019), 137--158.

\end{thebibliography}

% biblatex with biber
% \printbibliography                

%-------------------------------------------------------------------------

\end{document}